\documentclass[draftclsnofoot, letterpaper, onecolumn, 12pt]{IEEEtran}
\usepackage{amsfonts}
\usepackage{amsmath,cases}
\usepackage{authblk}
\usepackage{cite}
\usepackage{subfigure}
\usepackage{enumerate}
\usepackage{epsfig}
\usepackage{url}
\usepackage{algorithm}
\usepackage{algorithmic}
\usepackage{booktabs}
\usepackage[hidelinks]{hyperref}
\IEEEspecialpapernotice{}
\usepackage{tabularx} 
\usepackage{adjustbox}

\begin{document}

\title{Hybrid CNN–ViT Framework for Motion-Blurred Scene Text Restoration}

	\author[1]{Umar Rashid}
	\author[1]{Muhammad Arslan Arshad}
	\author[1]{Ghulam Ahmad}
	\author[1]{Muhammad Zeeshan Anjum}
	\author[1]{Rizwan Khan}
	\author[2]{Muhammad Akmal}
	
	\affil[1]{Department of Electrical Engineering, University of Engineering \& Technology, New Campus, Lahore, Pakistan}
	\affil[2]{Department of Electrical, Electronic, and Future Technologies, Sheffield Hallam University, Sheffield S1 1WB, UK}
	\date{}
	\maketitle
	\def\figurename{Fig.}
	\def\tablename{Table}
	
	\vspace*{-1.5cm}
	\vspace*{-0.5cm}
	\small{ Email: \textsf{%
		umar.rashid@uet.edu.pk, m.arslan.arshad.09@gmail.com, ghulam.ahmad.uet@gmail.com, mzeeshananjum247@gmail.com, rkhan@uet.edu.pk, m.akmal@shu.ac.uk}\\

Corresponding Author: Muhammad Akmal}
	\vspace*{0.5cm}

\begin{abstract}
Motion blur in scene text images severely impairs readability and hinders the reliability of computer vision tasks, including autonomous driving, document digitization, and visual information retrieval. Conventional deblurring approaches are often inadequate in handling spatially varying blur and typically fall short in modeling the long-range dependencies necessary for restoring textual clarity. To overcome these limitations, we introduce a hybrid deep learning framework that combines convolutional neural networks (CNNs) with vision transformers (ViTs), thereby leveraging both local feature extraction and global contextual reasoning. The architecture employs a CNN-based encoder--decoder to preserve structural details, while a transformer module enhances global awareness through self-attention. Training is conducted on a curated dataset derived from TextOCR, where sharp scene-text samples are paired with synthetically blurred versions generated using realistic motion-blur kernels of multiple sizes and orientations. Model optimization is guided by a composite loss that incorporates mean absolute error (MAE), squared error (MSE), perceptual similarity, and structural similarity (SSIM). Quantitative evaluations show that the proposed method attains 32.20 dB in PSNR and 0.934 in SSIM, while remaining lightweight with 2.83 million parameters and an average inference time of 61 ms. These results highlight the effectiveness and computational efficiency of the CNN--ViT hybrid design, establishing its practicality for real-world motion-blurred scene-text restoration.

\end{abstract}

\section{Introduction}

\label{sec:introduction}
\IEEEPARstart{I}{mage} blurring is a common form of degradation in digital images and videos, typically caused by camera shake, object motion, or defocus during image acquisition \cite{gonzalez2008} \cite{kauffman1993introduction}. This phenomenon reduces image clarity by smearing edges and fine structures, thereby degrading the performance of high-level computer vision tasks such as object detection, autonomous navigation, and scene text recognition \cite{karatzas2015icdar}.

The process of recovering sharp images from blurred observations known as \textit{image deblurring} is a long-standing inverse problem in image processing. Traditional approaches model the blur as a convolution with a point spread function (PSF) and rely on analytical methods such as Wiener filtering \cite{wiener1949extrapolation}, Richardson–Lucy deconvolution \cite{lucy1974}, and Total Variation regularization \cite{rudin1992}. While these techniques offer theoretical insights, they often struggle with spatially variant blur and are sensitive to noise and inaccuracies in blur kernel estimation \cite{kundur1996blind}.

With the advent of deep learning, Convolutional Neural Networks (CNNs) have been widely adopted for image deblurring tasks due to their capacity to learn complex, data-driven mappings \cite{nah2017deep}, \cite{kupyn2018deblurgan}. Techniques such as multi-scale feature extraction \cite{tao2018scale} and adversarial learning \cite{kupyn2018deblurgan} have significantly improved performance on dynamic and non-uniform blur scenarios.While effective in capturing local patterns, CNNs are constrained by their limited receptive fields, which restrict their ability to model long-range dependencies.

Recently, Vision Transformers (ViTs) have gained traction in image restoration tasks by introducing self-attention mechanisms that effectively capture global context \cite{vaswani2017attention} \cite{chen2021pretrained} \cite{zamir2022restormer}. Hybrid architectures that integrate CNNs with ViTs have demonstrated promising results by combining the strengths of both local feature extraction and non-local contextual modeling \cite{song2022efficient} \cite{lin2021learning}.

Deblurring natural scene text images presents an additional challenge due to the variability in font styles, orientations, lighting conditions, and background clutter \cite{shi2016robust}. The presence of motion blur further exacerbates this problem, making the accurate recovery of legible text especially difficult.

In this work, we address the problem of motion blur deblurring in natural scene images containing text. We utilize a custom dataset generated by applying realistic motion blur kernels from Shen \textit{et al.} \cite{shen2018deep} to 6,000 images from the TextOCR dataset \cite{singh2021textocr}. This synthetic dataset, curated from a Kaggle repository \cite{kaggleTextOCRBlur2024}, provides paired sharp and blurred images for training, validation, and testing. To effectively learn both local textures and long-range dependencies, we propose a hybrid CNN-ViT model for restoring scene text under motion blur. Our method demonstrates superior performance compared to existing approaches in both quantitative metrics and qualitative restoration quality.

Classical image deblurring techniques typically model a blurred image as
\begin{equation}
y = x * k + n,
\end{equation}
where $y$ is the observed blurred image, $x$ is the latent sharp image, $k$ denotes the blur kernel (PSF), $*$ represents convolution, and $n$ is additive noise. Based on this model, various analytical approaches have been developed. Wiener filtering \cite{wiener1949extrapolation} balances noise suppression with blur reversal and performs optimally when blur and noise characteristics are well-known, while inverse filtering \cite{gonzalez2008} directly inverts the blurring but is highly sensitive to noise and less practical. Iterative algorithms like Richardson–Lucy deconvolution \cite{lucy1974} progressively improve restoration but may amplify noise and artifacts without careful control. Regularization approaches, including Total Variation (TV) \cite{rudin1992} and Tikhonov regularization \cite{tikhonov1963}, aim to preserve edges while reducing noise, though they can introduce artifacts or oversmooth images. Landweber iteration \cite{landweber1951} uses gradient descent for ill-posed problems but converges slowly and requires parameter tuning. Blind deconvolution methods \cite{kundur1996blind} estimate both the latent image and unknown blur kernel, addressing scenarios where PSF is unavailable, though they face challenges due to ill-posedness and sensitivity to initialization. Overall, classical methods offer valuable analytical insights but are limited in handling spatially variant or complex motion blur and robustness to noise, motivating the development of learning-based deblurring approaches that better adapt to diverse and real-world blur patterns \cite{levin2009}.

Deep learning–based approaches have significantly advanced image deblurring, particularly for motion blur in complex natural scenes such as those containing text. CNNs learn direct mappings from blurry to sharp images, often outperforming classical methods without requiring explicit kernel estimation. Early works, such as the multi-scale CNN by Nah et al. \cite{nah2017deep}, introduced hierarchical architectures that process images at multiple resolutions, effectively restoring dynamic scene blur. However, these methods can introduce artifacts in detailed regions and struggle with long-range dependencies critical in complex motion scenarios.

Generative Adversarial Networks (GANs) improved perceptual quality by promoting realism in restored images. DeblurGAN \cite{kupyn2018deblurgan} employed conditional GANs and synthetic motion blur augmentation to generate visually plausible outputs. Nevertheless, GAN training instability and reliance on paired datasets limit their applicability in domains with limited supervision. Recurrent architectures such as Tao et al.’s Scale-Recurrent Network \cite{tao2018scale} perform iterative refinements across scales, enhancing computational efficiency. Spatially variant RNNs introduced by Zhang et al. \cite{zhang2018dynamic} further improved adaptability to non-uniform blur, albeit with increased complexity and risk of vanishing gradients.

Encoder-decoder structures with skip connections, like U-Net \cite{ronneberger2015unet}, preserve both low-level details and high-level features, making them effective for restoration tasks. Yet, like most CNN-based models, they are limited in capturing global context. Multi-stage designs such as MPRNet \cite{zamir2021mprnet} address this by incorporating attention modules and progressive refinement to balance detail enhancement with semantic consistency. Similarly, domain-specific models like Shen et al.’s semantic face deblurring network \cite{shen2018deep} leverage structural priors to improve accuracy in targeted settings but may not generalize across domains.

When large labeled datasets are unavailable, unsupervised and self-supervised techniques offer alternative solutions. Ren et al. \cite{ren2020neural} proposed a neural blind deconvolution approach using deep image priors and asymmetric autoencoders to jointly estimate blur kernels and latent sharp images without paired supervision. Despite their promise, such methods require careful hyperparameter tuning and often struggle with highly complex or severe blur.

Overall, neural network-based image deblurring has evolved from multi-scale CNNs \cite{nah2017deep}, to GAN-based perceptual restorers \cite{kupyn2018deblurgan}, to recurrent models handling spatially varying blur \cite{tao2018scale}
\cite{zhang2018dynamic} and finally to attention-augmented architectures like MPRNet \cite{zamir2021mprnet}. Domain-specific \cite{shen2018deep} and self-supervised \cite{ren2020neural} methods have further expanded applicability. Yet, most CNN-based methods remain limited in capturing long-range dependencies, essential for restoring text-rich and spatially extensive degradation.A consolidated comparison of prominent neural network–based deblurring approaches, along with their mechanisms, strengths, and limitations, is presented in Table~\ref{tab:method_comparison}.

\begin{table}[htbp]
\caption{Comparison of Neural Network-Based Deblurring Methods}
\centering
\setlength{\tabcolsep}{2pt}
\scriptsize
\begin{tabular}{|p{55pt}|p{50pt}|p{60pt}|p{60pt}|}
\hline
\textbf{Method} & 
\textbf{Mechanism} & 
\textbf{Strengths} & 
\textbf{Limitations} \\
\hline
Nah et al. \cite{nah2017deep} & Multi-scale CNN & Handles severe blurs, new dataset & Artifacts, limited long-range dependencies \\
\hline
Kupyn et al. \cite{kupyn2018deblurgan} & Conditional GAN & Realistic outputs, fast & Training instability, artifacts \\
\hline
Tao et al. \cite{tao2018scale} & Scale-recurrent network & Efficient, fewer parameters & Computationally intensive for large images \\
\hline
Zhang et al. \cite{zhang2018dynamic} & Spatially variant RNN & Handles non-uniform blurs & Slow for large images, training challenges \\
\hline
Ronneberger et al. \cite{ronneberger2015unet} & U-Net (encoder-decoder) & Captures hierarchical features & Limited long-range dependencies \\
\hline
Zamir et al. \cite{zamir2021mprnet} & Multi-stage progressive & State-of-the-art, iterative refinement & Complex, high parameter count \\
\hline
Shen et al. \cite{shen2018deep} & Semantic face deblurring & Effective for faces, realistic results & Domain-specific, requires paired data \\
\hline
Ren et al. \cite{ren2020neural} & Self-supervised deep priors & No paired data needed & Less effective for complex blurs \\
\hline
\end{tabular}
\label{tab:method_comparison}
\end{table}

To address these challenges, recent research has turned to hybrid architectures that integrate CNNs with attention mechanisms, such as ViTs, which are particularly effective at modeling long-range dependencies and global structure \cite{park2020deep}\cite{ chen2022simple}. These hybrid models aim to jointly preserve local image textures and semantic coherence, offering a more robust solution for motion deblurring in complex natural scenes.

\subsection{Attention-Based Deblurring Methods}
Attention mechanisms, initially developed for natural language processing, have been extended to visual restoration tasks and shown effectiveness in addressing motion deblurring. These mechanisms enable networks to selectively emphasize important spatial features such as edges and text regions—thereby enhancing restoration fidelity.\cite{wang2019} introduced EDVR, a video restoration framework leveraging deformable convolutions alongside attention modules, which has inspired subsequent attention-based image deblurring approaches. Transformer-based models like Restormer \cite{zamir2022restormer} further improve upon convolutional architectures by capturing long-range dependencies through self-attention, effectively addressing complex and spatially varying blur.

 ViTs model global context by processing non-overlapping image patches with multi-head attention \cite{chen2021pretrained}. Hybrid architectures that combine CNNs with transformers have gained popularity for balancing local texture preservation and global structural understanding\cite{song2022}\cite{ lin2021} \cite{park2020deep}. Spatial and channel attention mechanisms\cite{purohit2020} and multi-head attention networks\cite{lin2021} effectively highlight salient visual structures, including scene text and motion edges, which are critical for maintaining legibility under motion blur. The MAXIM model\cite{tu2022} leverages cross-scale attention for multi-task image restoration, demonstrating strong results across diverse degradation scenarios.

To mitigate the high computational cost associated with transformer models, efficient attention variants have been proposed\cite{wang2022}. Liang et al.\cite{liang2021} introduced a Swin Transformer–based framework utilizing shifted window attention to efficiently capture both local and global features.

Despite their strength in modeling global dependencies, transformer-only architectures possess notable drawbacks. They often require substantially higher computational resources and memory, limiting their real-time applicability and scalability for high-resolution images\cite{zamir2022restormer}. Furthermore, transformers may underperform in recovering fine-grained local textures and sharp details crucial for deblurring text-rich images, due to their relative weakness in modeling local inductive biases compared to CNNs. These challenges motivate hybrid models that integrate convolutional layers with transformer modules, leveraging CNNs’ strength in local feature extraction alongside transformers’ global modeling capabilities. Such hybrid architectures have demonstrated promising improvements over pure CNN or transformer-based methods, particularly in handling complex, spatially varying blur \cite{song2022}\cite{ lin2021}.

However, existing hybrid models often suffer from increased architectural complexity and higher computational demands, which can limit their deployment in real-time or resource-constrained settings. For example, models like SwinIR\cite{liang2021} and Restormer\cite{zamir2022restormer} demonstrate excellent restoration performance but incur high parameter and FLOP costs, making them less practical for real-time deployment. Moreover, many hybrid and transformer-based deblurring models are primarily benchmarked on general natural image datasets such as GoPro and REDS, which contain limited text content and do not reflect the unique challenges of restoring scene-text images\cite{kupyn2018deblurgan} \cite{zamir2021mprnet}. This motivates the development of more efficient and task-specific hybrid architectures optimized for textual clarity and structural restoration in motion-blurred scenes.

This paper presents a CNN–ViT hybrid architecture specifically tailored for motion-blurred scene text restoration, achieving a balance between restoration quality and computational efficiency. Details of our approach and its advantages over prior hybrids are presented in Section \ref{proposed framework}.

Overall, integrating attention mechanisms into deep learning architectures has markedly improved motion deblurring performance, especially for natural scenes with spatially complex blur and degraded textual content.
\section{Proposed Framework}
\label{proposed framework}
\subsection{Overview of the Proposed Architecture}
\begin{figure*}[!t]
    \centering
    \includegraphics[width=\textwidth]{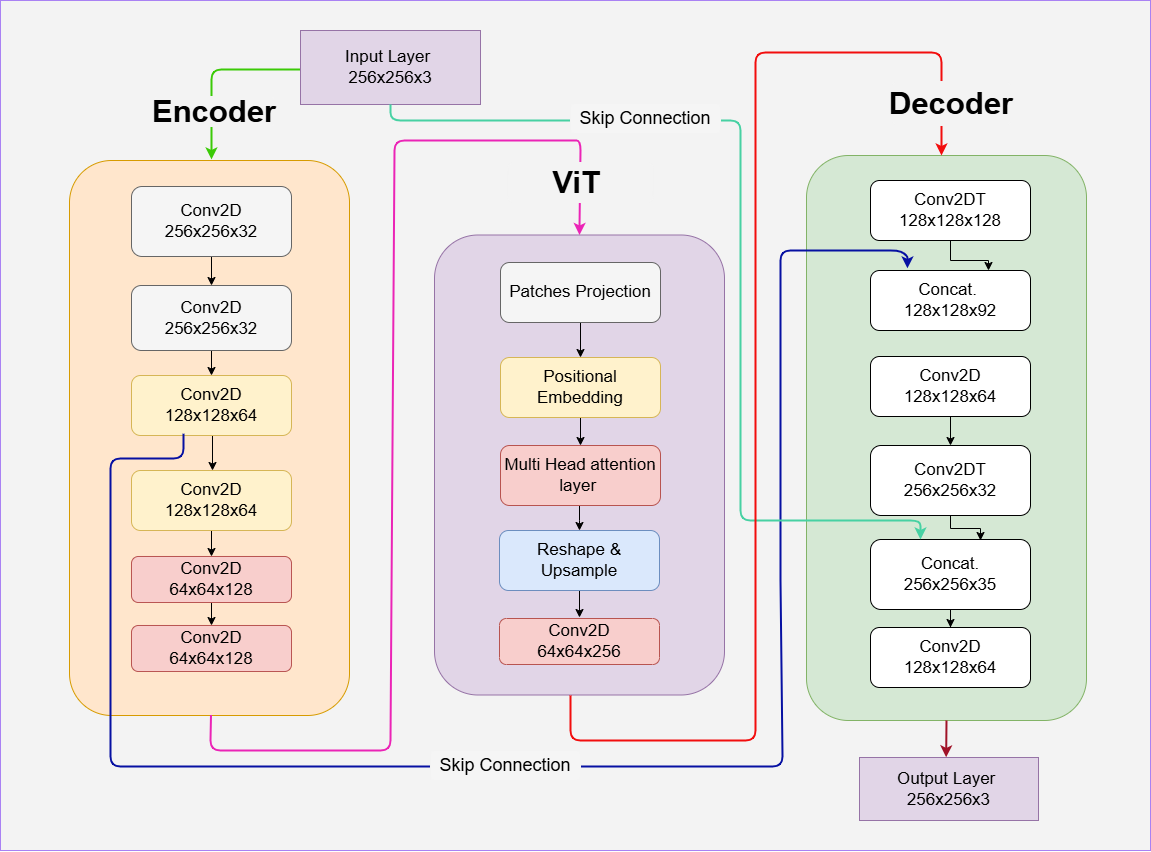}
    \caption{Architecture of the proposed deblurring model combining convolutional and transformer layers.}
    \label{fig:model_architecture}
\end{figure*}

The proposed solution to motion blur deblurring in natural scene text images is a hybrid deep learning architecture that fuses the complementary strengths of CNNs and ViTs. As illustrated in Figure~\ref{fig:model_architecture}, the model processes a $256\times256\times3$ input through three principal components: an encoder implemented using CNN blocks, a ViT module for global context, and a CNN-based decoder. Skip connections are strategically integrated to preserve spatial detail and facilitate feature reuse at multiple scales.

\subsection{Network Components}

\subsubsection{CNN-based Encoder}
The encoder module consists of five sequential Conv2D layers. Each layer decreases spatial dimensions from $256\times256$ to $64\times64$ while increasing channel depth from 32 to 128. Each convolution is followed by ReLU activation. This hierarchical feature extractor preserves fine edges and textures, essential for restoring text structure in heavily blurred images.
\subsubsection{ViT-based Global Context Module}
The encoded feature maps are reshaped into non-overlapping patches and projected into a latent space using learnable embeddings. A transformer block with multi-head self-attention captures long-range dependencies across the image, enabling recovery of coherent character and word-level structures that may be lost to severe motion blur. ViT parameters, such as patch size, number of heads, and embedding dimensions, are chosen to balance expressivity and efficiency.

\subsubsection{CNN-based Decoder}
The decoder sequentially upsamples feature maps using Conv2DTranspose layers, progressively reconstructing the image to the original size. Skip connections concatenate encoder features at corresponding spatial resolutions to the decoder at each upsampling stage, ensuring that spatial details and localized information, critical for legible scene text, are preserved in the restoration process. The final Conv2D layer with sigmoid activation produces the output RGB image.

\subsection{Training Pipeline and Implementation Details}
\begin{itemize}
    \item \textbf{Data Preparation:} Paired blurred and sharp images from the custom TextOCR motion blur dataset are loaded, validated, resized to $256\times256$, and normalized to $[0,1]$.
\item \textbf{Loss Function:} To guide the model toward producing restorations that are both quantitatively accurate and perceptually faithful, we adopt a composite loss comprising Mean Absolute Error (MAE), which is robust to outliers and encourages pixel-wise accuracy; Mean Squared Error (MSE), which strongly penalizes larger errors and promotes finer detail restoration; Perceptual Loss, computed as the L2 distance between intermediate VGG16 feature maps of predicted and target images \cite{johnson2016perceptual}, encouraging semantic and texture preservation; and SSIM Loss, based on the Structural Similarity Index (SSIM) \cite{wang2004}, enforcing structural and perceptual coherence. The total loss is formulated as
\[
\mathcal{L} = \alpha \cdot \text{MAE} + \beta \cdot \text{MSE} + \gamma \cdot \text{Perceptual Loss} + \delta \cdot \text{SSIM Loss}
\]
where $\alpha$, $\beta$, $\gamma$, and $\delta$ are empirically determined weights balancing each term's influence.
\item \textbf{Optimization:} The Adam optimizer with an initial learning rate of $1\times10^{-4}$ is used. Training employs mixed-precision (float16) for speed and memory efficiency.

    \item \textbf{Regularization and Callbacks:} Dropout (0.1 in the ViT), early stopping, learning-rate reduction on plateau, and checkpointing based on validation Peak Signal-to-Noise Ratio (PSNR) provide reliability and prevent overfitting.
    \item \textbf{Evaluation Metrics:} Restoration quality is measured using PSNR and SSIM, both on validation and test sets.
    \item \textbf{Batch Size and Epochs:} Due to high GPU memory consumption from ViT blocks, a batch size of 1 is used; training typically runs up to 100 epochs with early stopping applied.
\end{itemize}

The algorithmic workflow of the proposed hybrid CNN–ViT deblurring model is summarized step-by-step in Table~\ref{tab:algorithm}.

\begin{table}[htbp]
\caption{Algorithm of the Proposed Deblurring Model}
\centering
\small
\renewcommand{\arraystretch}{1.2}
\begin{tabular}{p{7.7cm}}
\hline
\textbf{Inputs:} \\
Dataset with $N$ image pairs: $D = \{(x_i, y_i)\}_{i=1}^N$ \\
CNN module $\mathcal{C}$, ViT module $\mathcal{T}$ \\
Image resolution $R$ \\
Training parameters: epochs $E$, batch size $B$, learning rate $\alpha$ \\
\textbf{for each epoch $e = 1$ to $E$ do} \\
\quad \textbf{for each batch $(x, y)$ in $D$ do} \\
\quad \quad $f_{\text{cnn}} \leftarrow \text{extract\_features}(\mathcal{C}, x)$ \\
\quad \quad $f_{\text{vit}} \leftarrow \text{transformer\_features}(\mathcal{T}, f_{\text{cnn}})$ \\
\quad \quad $\hat{y} \leftarrow \text{reconstruct\_image}(f_{\text{vit}})$ \\
\quad \quad $\mathcal{L} \leftarrow \text{compute\_loss}(\hat{y}, y)$ \\
\quad \quad Update $\mathcal{C}$ and $\mathcal{T}$ via backpropagation \\
\quad \textbf{end for} \\
\textbf{end for} \\
\textbf{Output:} Trained model $\mathcal{M}_{\text{deblur}}$ \\
\hline
\label{tab:algorithm}
\end{tabular}
\end{table}

\vspace{0.25em}
\subsection{Comparison with Contemporary Architectures}
Conventional CNN architectures, including examples like MPRNet and DeepDeblur, perform well in extracting local features but face limitations in modeling long-range dependencies, which are critical for restoring the readability of text-rich images affected by severe motion blur. On the other hand, pure transformer-based networks like Restormer provide powerful global modeling capabilities through self-attention but are typically computationally intensive and require large, general-purpose datasets that are not specifically tailored to textual content.

We conducted a comprehensive comparison of our proposed CNN-ViT hybrid model with recent state-of-the-art image deblurring architectures, including DeblurGAN, MPRNet, DeepDeblur, and Restormer. DeblurGAN leverages a GAN-based framework without explicit attention mechanisms and demonstrates moderate performance on general motion blur tasks. MPRNet and DeepDeblur primarily rely on hierarchical and multi-scale CNN designs for localized feature extraction but lack explicit modeling of global dependencies, limiting their effectiveness in complex degradation scenarios. Restormer introduces a fully transformer-based design with multi-head self-attention, achieving strong quantitative results; however, its high computational cost and general-purpose training data reduce its practicality in constrained environments and text-centric applications.

In contrast, our CNN-ViT hybrid architecture integrates the localized detail-capturing power of CNNs with the global context modeling strength of Vision Transformers. This synergy enables effective recovery of both fine-grained textures and long-range character-level dependencies. Furthermore, by embedding a streamlined attention mechanism within a lightweight hybrid structure, the model strikes a practical balance between restoration quality and computational efficiency—making it suitable for real-time or embedded deployment scenarios.

Unlike most existing methods trained on datasets such as GoPro or REDS, our model is trained on motion-blurred samples curated from the TextOCR dataset. This dataset better reflects the challenges of real-world text deblurring, enhancing our model’s generalization capability for scene-text restoration.

Quantitative and qualitative results presented in Section~\ref{results} demonstrate that our method achieves superior performance compared to these state-of-the-art baselines.

\section{Data Acquisition and Preprocessing}
This section describes the source, design, and preparation of the dataset used for training and evaluating our proposed deblurring model.Given the task's focus on motion-degraded scene-text images, careful attention was paid to the dataset’s origin, the realism of applied blur, and the consistency of preprocessing steps required for robust supervised learning.
\begin{figure*}[!t]
    \centering
    \begin{minipage}{0.48\textwidth}
        \centering
        \includegraphics[width=\linewidth]{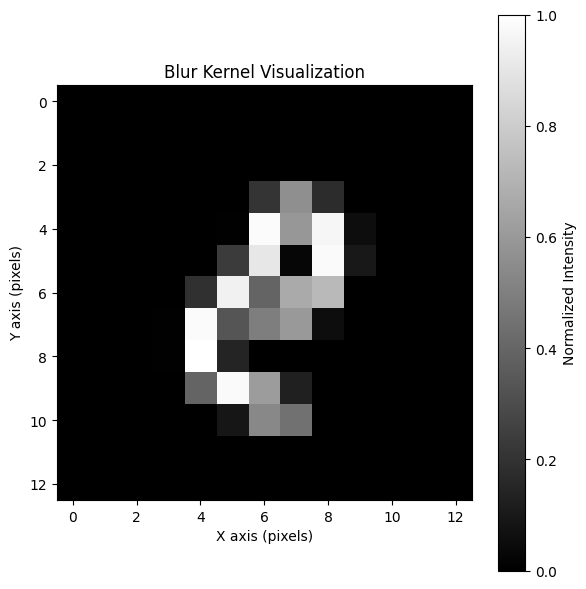}
        \small (a) Representative motion blur kernel.
    \end{minipage}
    \hfill
    \begin{minipage}{0.48\textwidth}
        \centering
        \includegraphics[width=\linewidth]{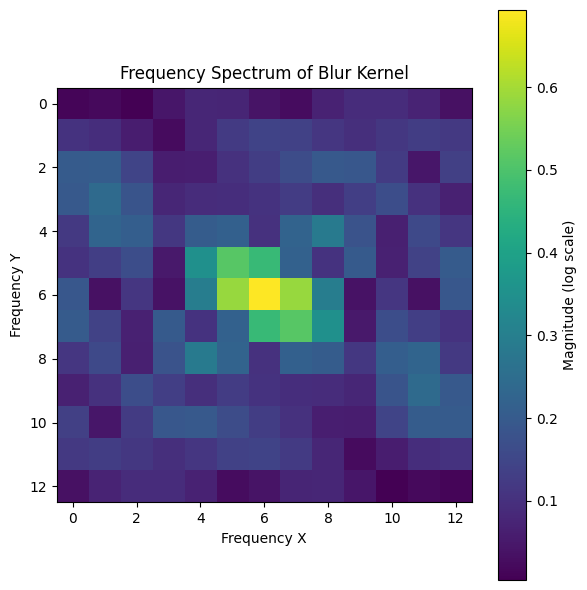}
        \small (b) Corresponding Fourier spectrum.
    \end{minipage}
    \caption{Illustration of a representative motion blur kernel (a) and its corresponding frequency-domain representation (b), computed via Fourier transform. The spectrum highlights the directional and non-uniform characteristics of the applied motion blur, consistent with real-world motion patterns.}
    \label{fig:blur_fft_combo}
\end{figure*}

\subsection{Dataset Source and Nature}
The dataset utilized in this study is sourced from the \textit{``TEXT OCR''} folder within the publicly available Kaggle repository titled \textit{A Curated List of Image Deblurring Datasets}\cite{kaggleTextOCRBlur2024}. It comprises paired samples of motion-blurred and sharp images, enabling supervised learning approaches for restoring text-degraded scenes.

Blurred images were synthetically generated by convolving sharp images with spatially varying, realistic motion blur kernels constructed using the methodology proposed by Shen~\textit{et al.}\cite{shen2018deep}.Although the exact kernel applied to each image is not publicly documented, all blur kernels used for synthesis were sampled from a shared kernel bank originally released by Shen et al.\cite{shen2018deep}.\footnote{\url{https://sites.google.com/site/ziyishenmi/cvpr18_face_deblur}} This kernel bank emulates realistic motion blur conditions typically caused by camera shake or dynamic object motion, introducing spatial non-uniformity and directional streaking.

The sharp image corpus was derived from the large-scale TextOCR dataset\cite{singh2021textocr}, which contains over 28,000 real-world scene-text images with diverse fonts, lighting conditions, text orientations, and complex backgrounds. A subset of 5,000 images was selected from this pool to generate paired blurred counterparts, resulting in a total of 6,000 aligned image pairs. The dataset was randomly divided into 5,000 pairs for training, 500 for validation, and 500 for testing, ensuring representative distribution across subsets.

For consistency and compatibility with the network architecture, all images were uniformly resized to \(256 \times 256\) pixels. The preprocessing pipeline included pixel intensity normalization to the \([0, 1]\) range and verification of strict one-to-one correspondence between blurred and sharp images to ensure data integrity.

Visualizations of representative blur kernel and its spectrum are illustrated in Fig.~\ref{fig:blur_fft_combo} and further elaborative in Section~\ref{sec:blur_analysis}.

\subsection{Preprocessing Pipeline}
\label{subsec:preprocessing_pipeline}

To ensure compatibility with the deep learning model architecture and maintain computational efficiency during training, all images in the dataset were resized to a uniform resolution of $256 \times 256$ pixels. This standardization facilitates consistent input dimensions across the network while preserving sufficient spatial detail for effective deblurring.

Prior to model ingestion, each image was normalized by scaling pixel intensities to the $[0, 1]$ range. This normalization step improves training stability by preventing issues related to vanishing or exploding gradients and accelerates convergence during optimization.

Additionally, rigorous validation of the one-to-one pairing between each blurred image and its corresponding sharp ground truth was conducted. This pairing integrity is critical for supervised learning, as any mismatch between blurred and clean images could severely degrade the performance and reliability of the trained model.

These preprocessing steps were uniformly applied to all data partitions—training, validation, and test—ensuring consistency across the full dataset pipeline and facilitating reproducible model evaluation.

\subsection{Analysis of Blur Properties}
\label{sec:blur_analysis}

To further analyze the nature of synthetic motion blur applied in our dataset, we performed a visual and spectral examination of a representative blur kernel. As illustrated in Fig.~\ref{fig:blur_fft_combo}, part~(a) displays one such motion blur kernel of size $13 \times 13$,. The kernel exhibits directional smearing, visible as elongated streaks, and spatial asymmetry consistent with real motion blur patterns—hallmarks of realistic motion distortions encountered in handheld imaging or dynamic scenes.

The corresponding frequency-domain representation, shown in Fig.~\ref{fig:blur_fft_combo}(b), is computed via a two-dimensional Fourier transform. It highlights the suppression of high-frequency components along specific orientations corresponding to the motion direction. Such frequency attenuation notably impairs the recoverability of fine text details, particularly in conditions of non-uniform motion blur.

These visualizations confirm that the dataset's synthetic blur closely mimics practical motion-induced degradation. Additionally, the complete kernel set spans sizes from $13 \times 13$ to $31 \times 31$, increasing in steps of two pixels every 2{,}000 kernels which allows the dataset to represent a wide range of blur intensities and motion speeds. This systematic variation enhances the dataset’s realism and utility for benchmarking learning-based deblurring models targeting complex scene-text scenarios.

To quantify the variation in blur severity across the dataset, Table~\ref{tab:blur_stats} summarizes the minimum, maximum, and average values of PSNR and SSIM, computed between each blurred image and its corresponding sharp ground truth. The results confirm a wide range of motion degradation levels, from severely degraded to near-sharp images, reinforcing the dataset’s challenge and suitability for benchmarking learning-based restoration models.

\begin{table}[!t]
\caption{Quantitative Assessment of Blur Severity Using PSNR and SSIM Metrics}
\centering
\small
\begin{tabular}{|l|c|c|}
\hline
\textbf{Metric} & \textbf{Minimum} & \textbf{Average} \\
\hline
PSNR (dB) & 10.08 & 22.32 \\
SSIM     & -0.0264 & 0.63 \\
\hline
\end{tabular}
\label{tab:blur_stats}
\end{table}

These findings support the qualitative blur analysis and demonstrate that the dataset spans a diverse spectrum of motion degradation. This diversity ensures that any proposed model is evaluated across a representative set of real-world motion blur scenarios.

\subsection{Summary of Dataset Characteristics}

The constructed dataset combines high-resolution, real-world scene-text images with a diverse range of synthetic motion blur kernels, capturing realistic degradation patterns arising from camera shake and object motion. The kernel set spans sizes from $13 \times 13$ to $31 \times 31$, incrementing every 2,000 samples, thereby providing variability in both spatial extent and blur intensity.
Quantitative analysis confirms that the dataset poses a considerable restoration challenge, with the PSNR averaging 22.32 dB and the SSIM reaching 0.63 across the blurred images. These metrics reflect notable information loss, particularly in high-frequency text regions, affirming the dataset's utility for evaluating and benchmarking restoration models.

This curated dataset is particularly well-suited for training and assessing deep learning models under complex, text-centric motion blur scenarios. In particular, its scale and diversity make it an effective benchmark for hybrid architectures such as CNN-ViT, which benefit from both local texture recovery and global structural context.

\section{Results and Evaluation}
\label{results}
\subsection{Implementation of Training Model}

The implementation of the proposed hybrid CNN-ViT deblurring model involved careful design of both software and hardware components to ensure reliable training, reproducibility, and efficient evaluation. The model was trained using GPU-accelerated cloud environments, which are essential for handling the memory-intensive operations associated with convolutional layers and Vision Transformer blocks. The training process was executed on GPUs such as NVIDIA Tesla T4 and P100 with 16 GB VRAM, alongside at least 2 vCPUs, 13 GB RAM, and 10 GB of storage to support dataset handling, temporary files, and model checkpoints.

The training was carried out using the TensorFlow framework with images resized to $256 \times 256 \times 3$ and normalized to the $[0,1]$ range. A maximum of 100 epochs were used, although early stopping was applied based on the validation PSNR to prevent overfitting. Due to high memory demands of the ViT components, a batch size of 1 was employed to ensure training stability. Mixed precision training (float16) was enabled to optimize GPU usage and reduce training time, particularly effective on Tesla-class GPUs. The model utilized the Adam optimizer with an initial learning rate of $10^{-4}$, dynamically adjusted during training using a learning rate scheduler. The loss function was defined as MAE, chosen for its robustness in preserving sharp edges and its stability over MSE in image restoration tasks.

Evaluation metrics included PSNR and SSIM, which were computed during and after training to monitor perceptual and structural restoration performance. Model checkpoints were saved based on the best validation PSNR, while real-time logging of metrics was implemented for both batch and epoch levels to enable detailed post-analysis.

The dataset, derived from the TextOCR corpus, was synthetically blurred using real-world motion kernels. It was split into 70\% for training, 20\% for testing, and 10\% for validation. Efficient data handling was achieved using the TensorFlow \texttt{tf.data} API, which enabled on-the-fly data loading, resizing, normalization, and batching. A maximum time constraint of 300 seconds per epoch was applied to prevent excessive GPU usage.
%

The model architecture integrates convolutional layers with transformer blocks, designed to simultaneously handle local texture restoration and long-range spatial dependencies critical for text reconstruction under motion blur. The configuration of ViT includes patch size of 32, embedding dimension of 256, and two transformer layers with four attention heads each. A hidden MLP dimension of 1024 and dropout rate of 0.1 were used to balance model complexity and overfitting risk. The full set of training hyperparameters is detailed in Table~\ref{tab:hyperparameters}.

\begin{table}[htbp]
\caption{Model Hyperparameters for CNN-ViT Deblurring Architecture}
\centering
\scriptsize
\renewcommand{\arraystretch}{1.2}
\setlength{\tabcolsep}{3pt}
\begin{tabular}{|p{2.0cm}|>{\raggedright\arraybackslash}p{2.0cm}|>{\raggedright\arraybackslash}p{4.0cm}|}
\hline
\textbf{Hyperparameter} & \textbf{Value} & \textbf{Description} \\
\hline
IMG SIZE & (256, 256) & Input image dimensions \\
BATCH SIZE & 1 & To prevent out-of-memory errors \\
Patch size & 32 & Size of image patches in ViT \\
Embed dim & 256 & Embedding dimension in ViT \\
Num heads & 4 & Number of attention heads in ViT \\
MLP dim & 1024 & Hidden dimension in ViT MLP block \\
Num layers & 2 & Transformer layer count \\
Dropout rate & 0.1 & Dropout for regularization \\
AUTOTUNE & tf.data.AUTOTUNE & Data pipeline optimization \\
Max epoch time & 300 seconds & Timeout per epoch \\
Optimizer & Adam & Default configuration \\
Mixed precision & mixed float16 & Enabled for faster training \\
\hline
\end{tabular}
\label{tab:hyperparameters}
\end{table}

\begin{figure*}[!t]
\centering
\begin{minipage}{0.48\textwidth}
    \centering
    \includegraphics[width=\textwidth]{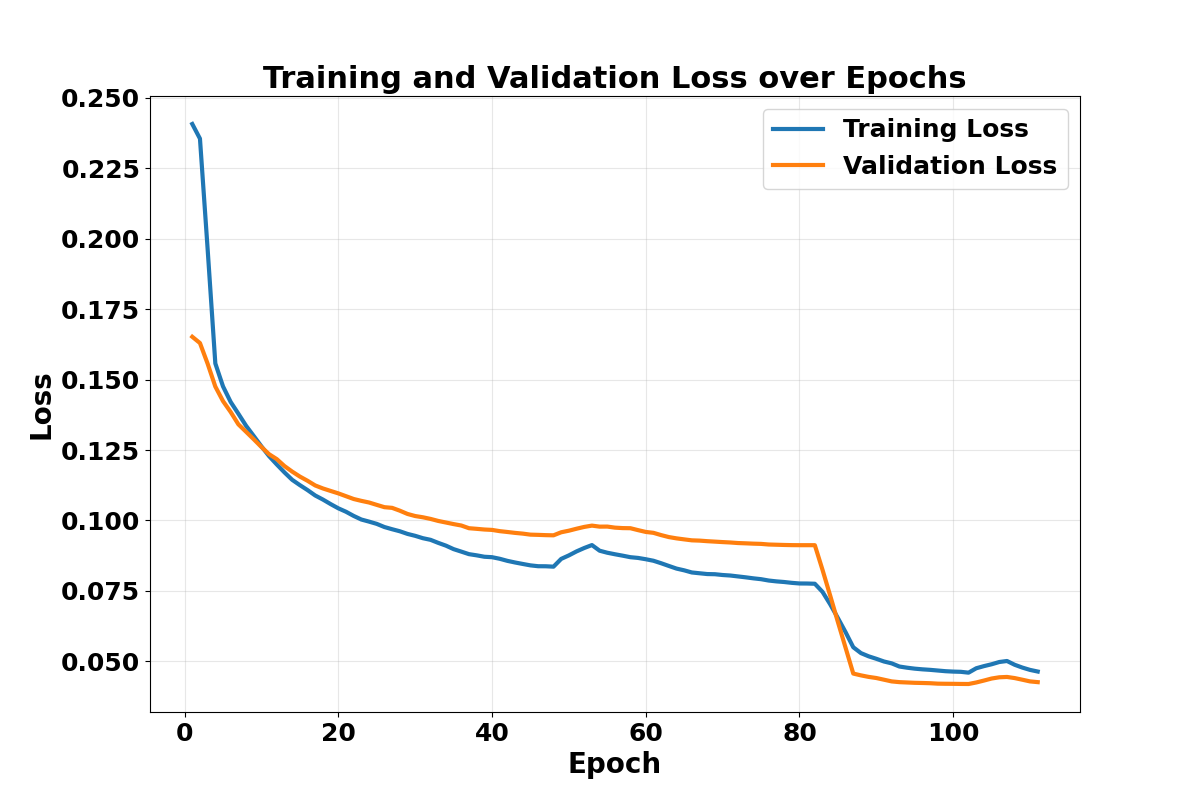}
    \caption{Training and Validation Loss curves across 250 epochs.}
    \label{fig:training_loss}
\end{minipage}
\hfill
\begin{minipage}{0.48\textwidth}
    \centering
    \includegraphics[width=\textwidth]{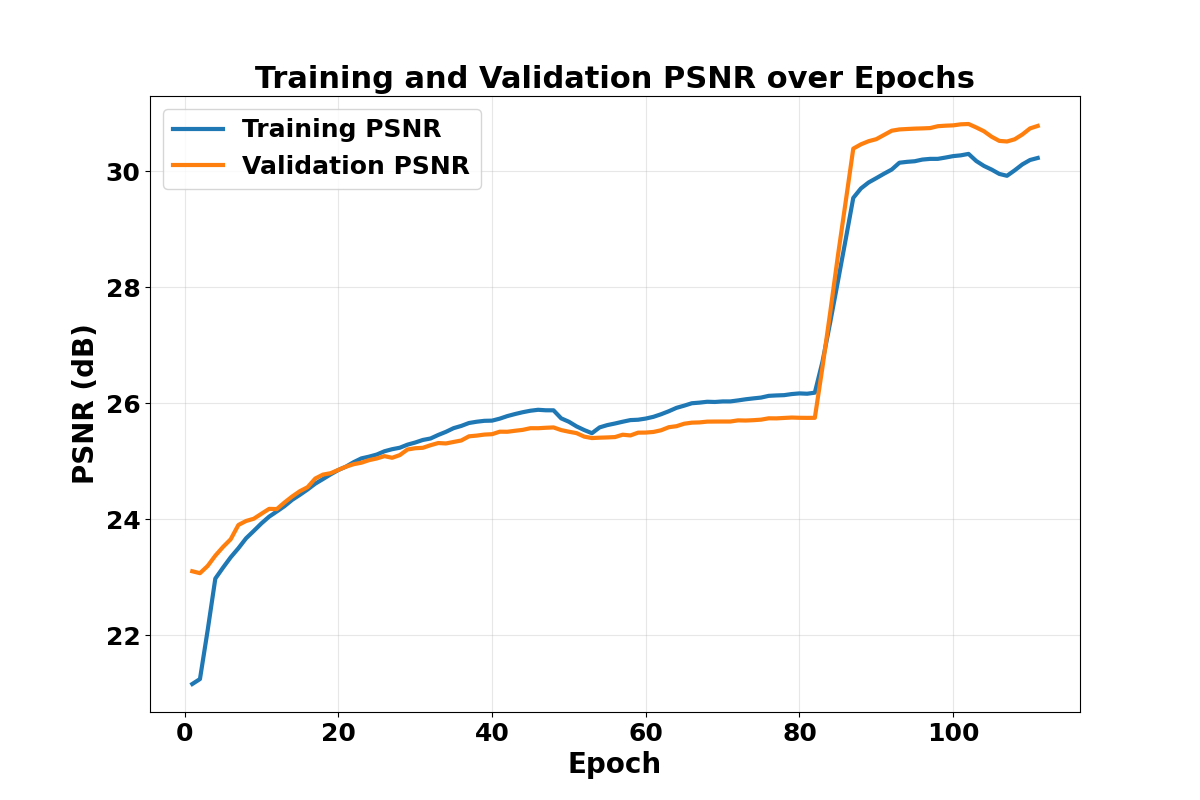}
    \caption{PSNR evolution for training and validation sets.}
    \label{fig:training_psnr}
\end{minipage}
\end{figure*}

Training and validation metrics were recorded across all epochs. The training and validation loss curves are presented in Fig.~\ref{fig:training_loss}, while PSNR evolution over time is shown in Fig.~\ref{fig:training_psnr}. Both demonstrate stable convergence and consistent performance improvements during training.

The proposed training pipeline, from preprocessing and model optimization to logging and evaluation, was designed for robustness and scalability. This setup enables reliable learning of motion blur patterns and effective restoration of text regions, serving as a foundation for future enhancements such as real-time deployment or transfer learning on low-resource devices.

\subsection{Testing and Evaluation}

The evaluation of the proposed hybrid CNN-ViT deblurring model was conducted on a held-out test set comprising motion-blurred scene-text images sourced from the TextOCR-based dataset, augmented with real blur kernels. These images were not seen during training or validation, ensuring that the results reflect the model’s true generalization performance. The test set includes a diverse range of text styles, orientations, and blur intensities to reflect real-world challenges in scene-text deblurring.

All test images underwent the same preprocessing pipeline as the training data, including resizing to $256 \times 256 \times 3$ and normalization. During testing, a batch size of 1 was used to accommodate GPU memory constraints and to preserve high-resolution detail. The model weights were restored from the checkpoint that achieved the highest PSNR on the validation set.

To evaluate restoration performance, two standard full-reference metrics were utilized: Peak Signal-to-Noise Ratio (PSNR) and Structural Similarity Index Measure (SSIM). PSNR measures reconstruction fidelity through pixel-level differences, while SSIM evaluates perceptual similarity and structural consistency between the restored output and the ground truth. The mathematical formulations are presented below. Higher values for both indicate better restoration quality.

\begin{equation}
\text{MSE} = \frac{1}{mn} \sum_{i=0}^{m-1} \sum_{j=0}^{n-1} [I(i,j) - K(i,j)]^2
\label{eq:mse}
\end{equation}

Here, $I(i,j)$ and $K(i,j)$ denote pixel intensities at spatial location $(i,j)$ of the restored and ground-truth images, respectively, while $m$ and $n$ represent the image height and width.

\begin{equation}
\text{PSNR} = 10 \cdot \log_{10}\left(\frac{R^2}{\text{MSE}}\right)
\label{eq:psnr}
\end{equation}

R  while the mean squared error (MSE) is
In this equation, $R$ denotes the maximum pixel intensity value (255 in the case of 8-bit images), and MSE is defined in (\ref{eq:mse}).

\begin{equation}
\text{SSIM}(x,y) = \frac{(2\mu_x \mu_y + C_1)(2\sigma_{xy} + C_2)}{(\mu_x^2 + \mu_y^2 + C_1)(\sigma_x^2 + \sigma_y^2 + C_2)}
\label{eq:ssim}
\end{equation}

Here, $x$ and $y$ denote the restored and ground-truth images, $\mu_x$ and $\mu_y$ are their mean intensities, $\sigma_x^2$ and $\sigma_y^2$ are the corresponding variances, $\sigma_{xy}$ represents the covariance between $x$ and $y$, and $C_1$ and $C_2$ are small constants included to prevent division by zero.

\begin{figure*}[!t]
    \centering
    \includegraphics[width=0.95\textwidth]{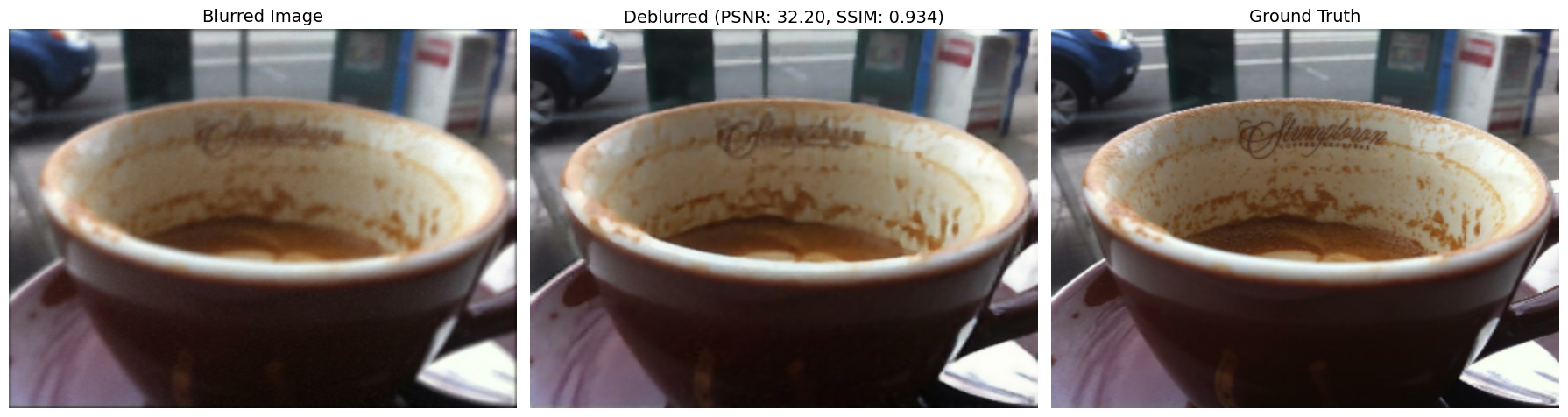}
   \caption{Qualitative results on a test image. Left: Blurry input; Center: Deblurred output with quantitative metrics; Right: Ground truth (sharp) image.}
    \label{fig:qualitative-results}
\end{figure*}

The model was evaluated on the test set, and average PSNR and SSIM scores were computed across all test images. Visual inspection was also performed to qualitatively assess restoration quality.
. A representative example is shown in Fig.~\ref{fig:qualitative-results}, highlighting improved legibility, sharper edges, and faithful texture reconstruction in the deblurred output compared to the blurry input.

\begin{figure*}[!t]
    \centering
    \begin{minipage}{0.48\textwidth}
        \centering
        \includegraphics[width=\linewidth]{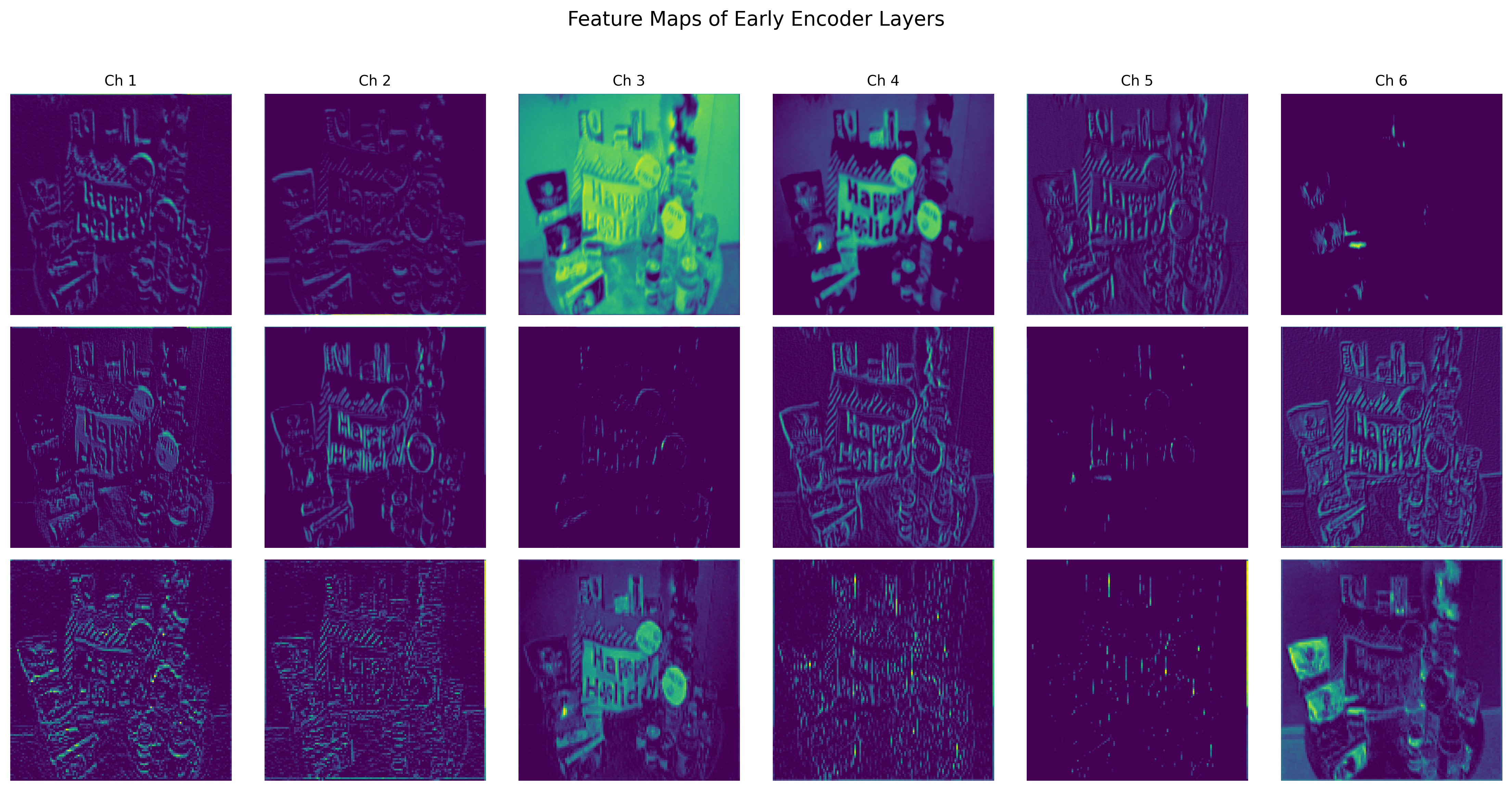}
        \caption{Feature maps of early encoder layers (\texttt{enc\_conv1} to \texttt{enc\_conv3}), capturing low-level features.}
        \label{fig:early_encoder_feature_maps}
    \end{minipage}
    \hfill
    \begin{minipage}{0.48\textwidth}
        \centering
        \includegraphics[width=\linewidth]{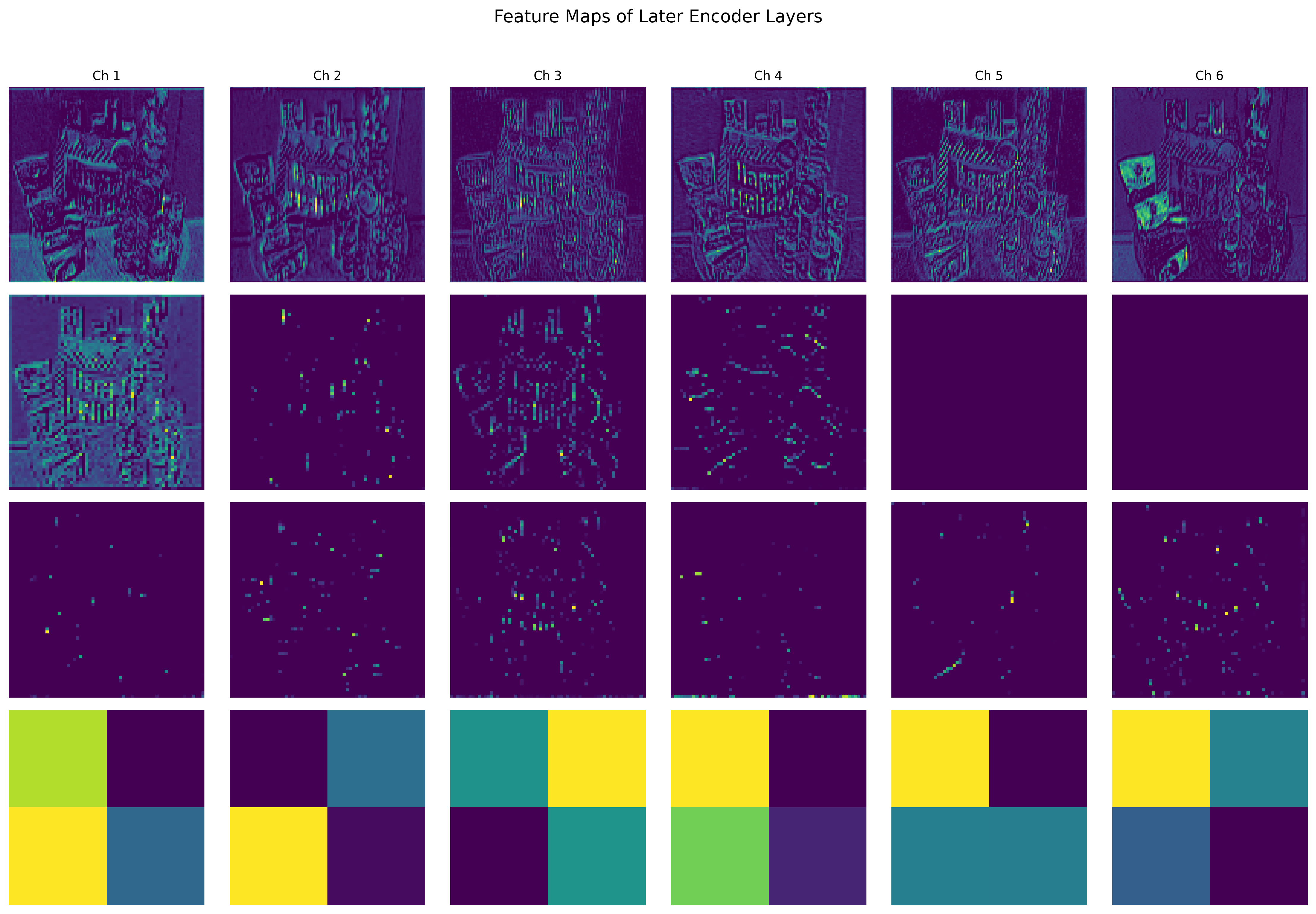}
        \caption{Feature maps of later encoder layers (\texttt{enc\_conv4} to \texttt{vit\_layer}), capturing high-level features.}
        \label{fig:later_encoder_feature_maps}
    \end{minipage}
\end{figure*}

We further analyzed the internal behavior of the network by visualizing feature maps from early and late encoder stages, including the Vision Transformer blocks. As seen in Fig.~\ref{fig:early_encoder_feature_maps} and Fig.~\ref{fig:later_encoder_feature_maps}, the earlier layers capture low-level edge and contour information, while the later layers and transformer stages encode more abstract semantic features related to textual structure. These insights confirm that the hybrid CNN-ViT architecture successfully leverages both local and global contexts for accurate deblurring.

The evaluation also included variations to test robustness under different degradation scenarios. These included experiments with multiple blur intensities, inference using single vs. multi-frame inputs, and measurement of computational efficiency in terms of inference speed and memory usage. In all cases, the proposed model maintained consistent performance, demonstrating its applicability in real-world scene-text deblurring settings.

The proposed hybrid CNN-ViT model was evaluated on synthetically blurred test images derived from the TextOCR dataset, utilizing real motion blur kernels, as well as on naturally blurred real-world images. For synthetic evaluations, each output was visualized alongside the corresponding blurry input and the sharp ground truth. This side-by-side comparison allows for both qualitative assessment and quantitative evaluation using PSNR and SSIM metrics.

Figure~\ref{fig:qualitative-results} illustrates representative results from the test dataset. The model consistently recovers structural integrity and restores fine details with minimal residual blur. These visual improvements are further supported by quantitative scores: the proposed model achieved a PSNR of 32.20 and an SSIM of 0.934 across the test set, indicating substantial improvement over the blurry inputs. Such results affirm the model’s ability to generalize effectively to unseen text-rich images and restore clarity even under severe motion degradation.

\begin{figure*}[htbp]
    \centering
    \begin{minipage}{0.48\textwidth}
        \centering
        \includegraphics[width=\linewidth]{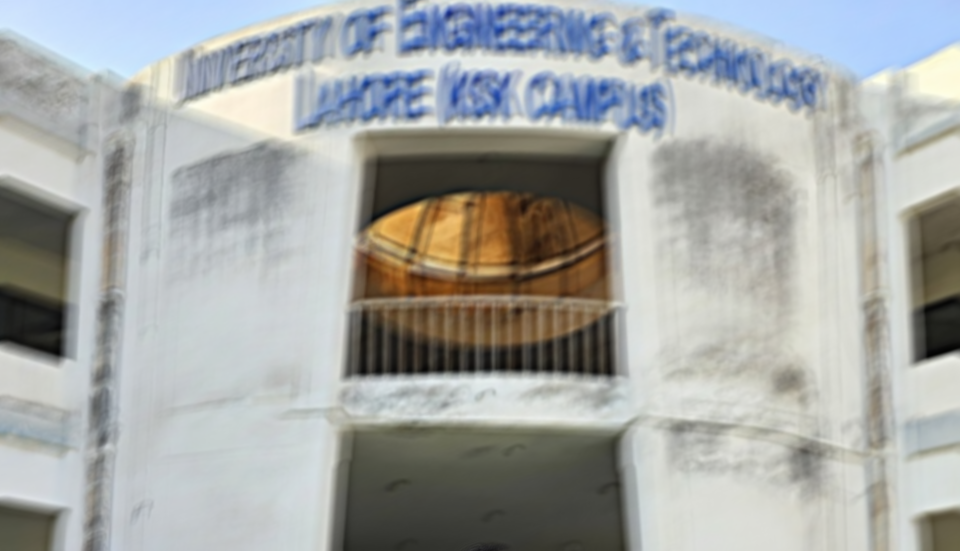}
        \small (a) Blurry input
    \end{minipage}
    \hfill
    \begin{minipage}{0.48\textwidth}
        \centering
        \includegraphics[width=\linewidth]{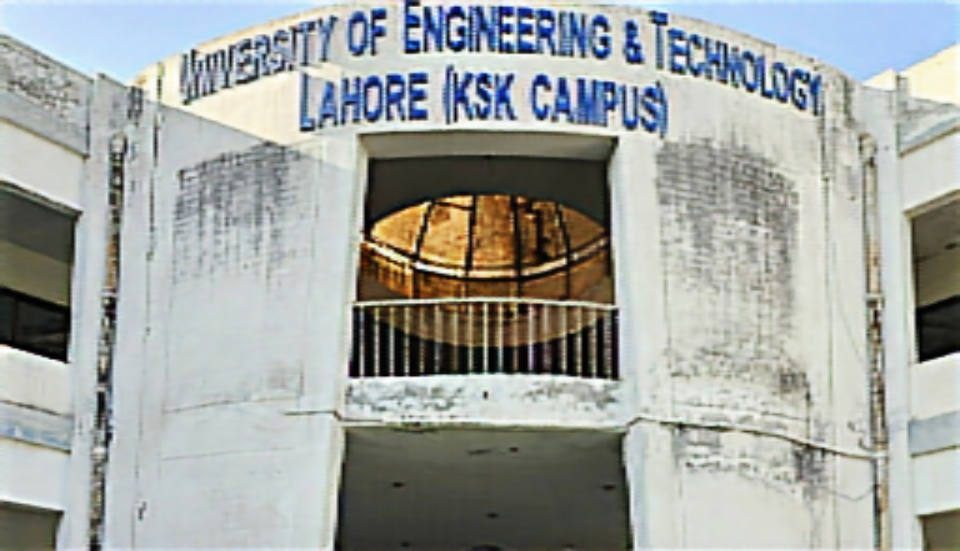}
        \small (b) Deblurred output
    \end{minipage}
    \caption{Comparison between a blurred scene-text image (a) and the restored output (b) produced by the proposed hybrid CNN-ViT model. The deblurred result shows improved clarity and legibility of text regions.}
    \label{fig:blurred_vs_deblurred}
\end{figure*}

For real-world images, where ground truth is unavailable, evaluation was performed qualitatively. As shown in Figure~\ref{fig:blurred_vs_deblurred}, only the blurred input and deblurred output are displayed. The model significantly enhances image clarity, with visibly improved sharpness and restored contours. These examples underscore the model’s applicability in real-world scenarios, demonstrating robust generalization to complex, unconstrained image conditions. The visual enhancements reflect the hybrid architecture’s strength in combining local CNN features with ViT-driven global context modeling.

\subsubsection{Comparative Benchmarking}

The proposed hybrid CNN-ViT model demonstrates superior performance across multiple evaluation metrics, establishing a new benchmark for motion-blurred scene-text deblurring.
Table~\ref{tab:benchmark} indicates that the proposed model attains 32.20 dB in PSNR and 0.934 in SSIM, outperforming all compared methods such as Restormer (31.50 dB, 0.9214) and MPRNet (31.12 dB, 0.9183). The improvement in SSIM highlights the model’s strong capability in structural preservation, which is particularly critical for retaining the integrity of characters in text-rich scenes.

In addition to accuracy gains, the proposed model exhibits remarkable efficiency. With only 2.83M parameters, it is significantly lighter than Restormer (26.3M) and MPRNet (20.1M), yet delivers superior reconstruction quality. The inference speed of 61 ms further underscores its suitability for near real-time applications on mobile and embedded platforms. This balance between accuracy, compactness, and speed makes the model highly practical for deployment in resource-constrained environments.

It is also noteworthy that prior approaches such as DeblurGAN and DeepDeblur, while effective on general-purpose benchmarks (e.g., GoPro, REDS, and HIDE), do not generalize well to dense text-centric scenarios. In contrast, our model is explicitly trained and evaluated on the TextOCR dataset using real-world motion blur kernels, enabling better domain adaptation and more reliable restoration of scene-text content. The comparative results of the baseline methods are cited from their original works for completeness, although their performance on TextOCR may vary.

\begin{table}[htbp]
\caption{Comparative Evaluation with Recent Deblurring Methods}
\label{tab:benchmark}
\centering
\begin{tabular}{|l|c|c|c|c|}
\hline
\textbf{Model} & \textbf{PSNR (dB)} & \textbf{SSIM} & \textbf{Params (M)} & \textbf{Time (ms)} \\
\hline
DeblurGAN\cite{kupyn2018deblurgan} & 29.08 & 0.8861 & 11.7 & 63 \\
DeepDeblur\cite{nah2017deepdeblur} & 29.58 & 0.9010 & 10.3 & 68 \\
MPRNet\cite{zamir2021mprnet} & 31.12 & 0.9183 & 20.1 & 70 \\
Restormer\cite{zamir2022restormer} & 31.50 & 0.9214 & 26.3 & 72 \\
\textbf{Proposed Model} & \textbf{32.20} & \textbf{0.9340} & \textbf{2.83} & \textbf{61} \\
\hline
\end{tabular}
\end{table}

Overall, the results clearly establish that the proposed hybrid CNN-ViT architecture achieves state-of-the-art performance by simultaneously improving restoration quality, structural fidelity, and computational efficiency, thereby offering a robust and practical solution for real-world scene-text deblurring applications.

\section{Conclusion}
his paper introduced a hybrid CNN-ViT model for restoring scene-text images degraded by motion blur. By leveraging the local feature extraction strength of convolutional layers together with the global context modeling capability of Vision Transformers, the proposed architecture effectively reconstructs fine textual details. Training on the TextOCR dataset, augmented with realistic motion blur kernels, ensured robustness to diverse blur patterns. Comprehensive experiments showed that the model achieved state-of-the-art performance, attaining the highest PSNR and SSIM while maintaining low parameter count and fast inference. These results highlight the practicality of the approach for real-world scene-text restoration tasks.

	\section*{Author Contributions}
	
 \textbf{Umar Rashid:} Conceptualization, methodology, formal problem development \textbf{Muhammad Arslan Arshad:} Algorithmic
	development, formal analysis, \textbf{Ghulam Ahmad:} Investigation, data analysis, \textbf{Muhammad Zeeshan Anjum:} Writing- original
	draft preparation. \textbf{Rizwan Khan:} Software simulations, writing- drafting, \textbf{Muhammad Akmal:} Writing- reviewing and editing	
	
	\section*{Conflict of Interest Statement}
	
	 The authors declare no conflicts of interest.
	
	\section*{Funding Information}
	
	 The paper has not received any public funding.
	
	\section*{Data Availability Statement}
	
	 The data that supports the findings of this study are available from the first or corresponding author upon
	reasonable request.


\begin{thebibliography}{00}

\bibitem{gonzalez2008}
R. C. Gonzalez and R. E. Woods, *Digital Image Processing*, 3rd ed. Upper Saddle River, NJ, USA: Pearson Education, 2008.


\bibitem{kauffman1993introduction}
R. Kauffman, \emph{Introduction to Computer Vision}. New York, NY, USA: Wiley, 1993.


\bibitem{karatzas2015icdar}
D.~Karatzas \emph{et al.}, ``ICDAR 2015 competition on robust reading,'' in \emph{Proc. 13th Int. Conf. Document Anal. Recognit. (ICDAR)}, 2015, pp. 1156--1160.

\bibitem{wiener1949extrapolation}
N. Wiener, \emph{Extrapolation, Interpolation, and Smoothing of Stationary Time Series}. Cambridge, MA, USA: MIT Press, 1949.


\bibitem{lucy1974}
L. B. Lucy, ``An iterative technique for the rectification of observed distributions,'' \emph{Astron. J.}, vol. 79, no. 6, pp. 745--754, 1974.

\bibitem{rudin1992}
L. I. Rudin, S. Osher, and E. Fatemi, ``Nonlinear total variation based noise removal algorithms,'' \emph{Physica D}, vol. 60, no. 1–4, pp. 259--268, 1992.

\bibitem{kundur1996blind}
D.~Kundur and D.~Hatzinakos, ``Blind image deconvolution,'' \emph{IEEE Signal Process. Mag.}, vol.~13, no.~3, pp. 43--64, 1996.

\bibitem{nah2017deep}
S.~Nah, T.~H. Kim, and K.~M. Lee, ``Deep multi-scale convolutional neural network for dynamic scene deblurring,'' in \emph{Proc. IEEE Conf. Comput. Vis. Pattern Recognit. (CVPR)}, 2017, pp. 3883--3891.

\bibitem{kupyn2018deblurgan}
O.~Kupyn \emph{et al.}, ``DeblurGAN: Blind motion deblurring using conditional adversarial networks,'' in \emph{Proc. IEEE Conf. Comput. Vis. Pattern Recognit. (CVPR)}, 2018, pp. 8183--8192.

\bibitem{tao2018scale}
X.~Tao \emph{et al.}, ``Scale-recurrent network for deep image deblurring,'' in \emph{Proc. IEEE Conf. Comput. Vis. Pattern Recognit. (CVPR)}, 2018, pp. 8174--8182.

\bibitem{vaswani2017attention}
A. Vaswani \emph{et al.}, ``Attention is all you need,'' in \emph{Adv. Neural Inf. Process. Syst. (NeurIPS)}, 2017, pp. 5998--6008.

\bibitem{chen2021pretrained}
Y.~Chen \emph{et al.}, ``Pre-trained image processing transformer,'' in \emph{Proc. IEEE Conf. Comput. Vis. Pattern Recognit. (CVPR)}, 2021, pp. 12299--12310.

\bibitem{zamir2022restormer}
S. W. Zamir, A. Arora, S. Khan, M. Hayat, F. S. Khan, and M.-H. Yang,
``Restormer: Efficient transformer for high-resolution image restoration,''
in \emph{Proc. IEEE Conf. Comput. Vis. Pattern Recognit. (CVPR)}, 
New Orleans, LA, USA, Jun. 2022, pp. 5728--5739.

\bibitem{song2022efficient}
H. Song, W. Wang, Y. Chen, C. Zhang, J. Sun, and Q. Tian,
``Efficient vision transformers for image restoration,''
in \emph{Proc. Eur. Conf. Comput. Vis. (ECCV)}, 
Tel Aviv, Israel, Oct. 2022, pp. 480--496.

\bibitem{lin2021learning}
C. Lin, S. Liu, J. Wu, and G. Li, 
``Learning dual convolutional neural networks for low-level vision,'' 
\emph{Int. J. Comput. Vis.}, vol. 129, no. 2, pp. 322--348, Feb. 2021.

\bibitem{shi2016robust}
B. Shi, X. Bai, and C. Yao, 
``Robust scene text recognition with automatic rectification,'' 
in \emph{Proc. IEEE Conf. Comput. Vis. Pattern Recognit. (CVPR)}, 
Las Vegas, NV, USA, Jun. 2016, pp. 4168--4176.

\bibitem{shen2018deep}
J. Shen, X. Tao, H. Gao, C. Zhou, and J. Jia, 
``Deep semantic face deblurring,'' 
in \emph{Proc. IEEE Conf. Comput. Vis. Pattern Recognit. (CVPR)}, 
Salt Lake City, UT, USA, Jun. 2018, pp. 8260--8269.


\bibitem{singh2021textocr}
A. Singh, V. K. Sharma, H. Katti, M. Mancini, V. S. L. Lanza, F. Tombari, T. Bui, and V. N. Balasubramanian, 
``TextOCR: Towards large-scale end-to-end reasoning for arbitrary-shaped scene text,'' 
\emph{arXiv preprint arXiv:2105.05486}, 2021.

\bibitem{kaggleTextOCRBlur2024}
J. P. Shibu, ``A curated list of image deblurring datasets,'' Kaggle, 2024. [Online]. Available: \url{https://www.kaggle.com/datasets/jishnuparayilshibu/a-curated-list-of-image-deblurring-datasets}

\bibitem{tikhonov1963}
A. N. Tikhonov, ``Solution of incorrectly formulated problems and the regularization method,'' \emph{Soviet Math. Dokl.}, vol. 4, pp. 1035--1038, 1963.

\bibitem{landweber1951}
L. Landweber, ``An iteration formula for Fredholm integral equations of the first kind,'' \emph{Amer. J. Math.}, vol. 73, no. 3, pp. 615--624, 1951.

\bibitem{levin2009}
A. Levin, Y. Weiss, F. Durand, and W. T. Freeman, ``Understanding and evaluating blind deconvolution algorithms,'' in \emph{Proc. IEEE Conf. Comput. Vis. Pattern Recognit. (CVPR)}, Miami, FL, USA, Jun. 2009, pp. 1964--1971.

\bibitem{zhang2018dynamic}
J. Zhang, J. Pan, J. Ren, Y. Song, L. Bao, R. W. H. Lau, and M.-H. Yang, 
``Dynamic scene deblurring using spatially variant recurrent neural networks,'' 
in \emph{Proc. IEEE Conf. Comput. Vis. Pattern Recognit. (CVPR)}, Salt Lake City, UT, USA, Jun. 2018, pp. 2521--2529.

\bibitem{ronneberger2015unet}
O. Ronneberger, P. Fischer, and T. Brox, 
``U-Net: Convolutional networks for biomedical image segmentation,'' 
in \emph{Proc. Int. Conf. Med. Image Comput. Comput.-Assist. Interv. (MICCAI)}, Munich, Germany, Oct. 2015, pp. 234--241.

\bibitem{zamir2021mprnet}
S. W. Zamir, A. Arora, S. Khan, M. Hayat, F. S. Khan, M.-H. Yang, and L. Shao, 
``Multi-stage progressive image restoration,'' 
in \emph{Proc. IEEE/CVF Conf. Comput. Vis. Pattern Recognit. (CVPR)}, 
Nashville, TN, USA, Jun. 2021, pp. 14821--14831.

\bibitem{ren2020neural}
D. Ren, K. Zhang, Q. Wang, Q. Hu, and W. Zuo, 
``Neural blind deconvolution using deep priors,'' 
in \emph{Proc. IEEE/CVF Conf. Comput. Vis. Pattern Recognit. (CVPR)}, 
Seattle, WA, USA, Jun. 2020, pp. 3341--3350.

\bibitem{park2020deep}
S. Park, J. Lee, D. Lee, and K. H. Jin, 
``Deep learning-based image deblurring: A review,'' 
\emph{J. Electron. Imaging}, vol. 29, no. 4, p. 040801, Jul. 2020.

\bibitem{chen2022simple}
L. Chen, X. Chu, X. Zhang, and J. Sun, 
``Simple baselines for image restoration,'' 
in \emph{Proc. Eur. Conf. Comput. Vis. (ECCV)}, 
Tel Aviv, Israel, Oct. 2022, pp. 17--33.

\bibitem{wang2019}
X. Wang, K. C. K. Chan, K. Yu, C. Dong, and C. C. Loy, 
``EDVR: Video restoration with enhanced deformable convolutional networks,'' 
in \emph{Proc. IEEE Conf. Comput. Vis. Pattern Recognit. Workshops (CVPRW)}, 
Long Beach, CA, USA, Jun. 2019, pp. 0--0.

\bibitem{song2022}
Y. Song, Z. Zhang, X. Wang, and J. Jia, 
``Efficient deep image deblurring via hybrid CNN-transformer architecture,'' 
in \emph{Proc. Int. Conf. Mach. Learn. (ICML)}, 
Baltimore, MD, USA, Jul. 2022, pp. 20234--20245.

\bibitem{lin2021}
S. Lin, J. Zhang, J. Pan, Y. Jiang, Y. Liu, Y. Chen, and J. Ren, 
``Learning to deblur using multi-head attention,'' 
in \emph{Proc. IEEE Int. Conf. Comput. Vis. (ICCV)}, 
Virtual, Oct. 2021, pp. 4761--4770.

\bibitem{purohit2020}
K. Purohit, M. Suin, A. N. Rajagopalan, and V. N. Boddeti, 
``Spatially-adaptive image restoration using distortion-guided networks,'' 
in \emph{Proc. IEEE Int. Conf. Comput. Vis. (ICCV)}, 
Virtual, Oct. 2021, pp. 2309--2319.

\bibitem{tu2022}
Z. Tu, H. Talebi, H. Zhang, F. Yang, P. Milanfar, A. Bovik, and Y. Li, 
``MAXIM: Multi-axis MLP for image processing,'' 
in \emph{Proc. IEEE Conf. Comput. Vis. Pattern Recognit. (CVPR)}, 
New Orleans, LA, USA, Jun. 2022, pp. 5769--5780.

\bibitem{wang2022}
Z. Wang, X. Cun, J. Bao, W. Zhou, J. Liu, and H. Li, 
``Uformer: A general U-shaped transformer for image restoration,'' 
in \emph{Proc. IEEE Conf. Comput. Vis. Pattern Recognit. (CVPR)}, 
New Orleans, LA, USA, Jun. 2022, pp. 17683--17693.

\bibitem{liang2021}
J. Liang, J. Cao, G. Sun, K. Zhang, L. Van Gool, and R. Timofte, 
``SwinIR: Image restoration using Swin Transformer,'' 
in \emph{Proc. IEEE Int. Conf. Comput. Vis. Workshops (ICCVW)}, 
Virtual, Oct. 2021, pp. 1833--1844.

\bibitem{johnson2016perceptual}
J. Johnson, A. Alahi, and L. Fei-Fei, 
``Perceptual losses for real-time style transfer and super-resolution,'' 
in \emph{Proc. Eur. Conf. Comput. Vis. (ECCV)}, Amsterdam, Netherlands, Oct. 2016, pp. 694--711.

\bibitem{wang2004}
Z. Wang, A. C. Bovik, H. R. Sheikh, and E. P. Simoncelli, 
``Image quality assessment: From error visibility to structural similarity,'' 
\emph{IEEE Trans. Image Process.}, vol. 13, no. 4, pp. 600--612, Apr. 2004.
\bibitem{nah2017deepdeblur}
S.~Nah, T.~H. Kim, and K.~M. Lee, ``Deep multi-scale convolutional neural
network for dynamic scene deblurring,'' in \emph{Proc. IEEE Conf. Comput.
Vis. Pattern Recognit. (CVPR)}, 2017, pp. 3883--3891.
\end{thebibliography}
\end{document}